# Single-Round Clustered Federated Learning via Data Collaboration Analysis for Non-IID Data

**Sota Sugawara[1], Yuji Kawamata[2,3], Akihiro Toyoda[1], Tomoru Nakayama[1], Yukihiko Okada[2,3]**
[1]Graduate School of Science and Technology, University of Tsukuba, Tsukuba, Japan
[2]Center for Artificial Intelligence Research, Tsukuba Institute for Advanced Research, University of Tsukuba, Tsukuba, Japan
[3]Institute of Systems and Information Engineering, University of Tsukuba, Tsukuba, Japan
yjkawamata@gmail.com

## Abstract

Federated Learning (FL) enables distributed learning across multiple clients without sharing raw data. When statistical heterogeneity across clients is severe, Clustered Federated Learning (CFL) can improve performance by grouping similar clients and training cluster-wise models. However, most CFL approaches rely on multiple communication rounds for cluster estimation and model updates, which limits their practicality under tight constraints on communication rounds. We propose Data Collaboration-based Clustered Federated Learning (DC-CFL), a single-round framework that completes both client clustering and cluster-wise learning, using only the information shared in DC analysis. DC-CFL quantifies inter-client similarity via total variation distance between label distributions, estimates clusters using hierarchical clustering, and performs cluster-wise learning via DC analysis. Experiments on multiple open datasets under representative non-IID conditions show that DC-CFL achieves accuracy comparable to multi-round baselines while requiring only one communication round. These results indicate that DC-CFL is a practical alternative for collaborative AI model development when multiple communication rounds are impractical. Our source code is publicly available at https://github.com/souta-suga/DC-CFL.

## 1 Introduction

In healthcare and finance, confidential (i.e., sensitive) data are distributed across clients, making it difficult to aggregate raw data at a single site for centralized training and analysis. Specifically, constraints related to legal compliance, information security, and organizational consensus building often prevent centralized learning that requires sharing raw data. In such settings, Federated Learning (FL) [McMahan *et al.*, 2017] has been widely studied as a framework for integrating knowledge from multiple clients without sharing raw data.

However, non-independent and identically distributed (non-IID) conditions are common in practical deployments. When data distributions differ across clients, inconsistent updates can prevent a single global model from generalizing sufficiently, leading to degraded performance [Li *et al.*, 2020; Zhu *et al.*, 2021]. Therefore, maintaining the benefits of collaborative learning while adapting to non-IID data remains an important challenge [Kairouz *et al.*, 2021].

As a promising approach to handling non-IID data, Clustered Federated Learning (CFL) has been proposed to enable collaborative learning among clients with similar data distributions [Sattler *et al.*, 2020]. However, many CFL methods rely on multiple communication rounds for cluster estimation and model updates, which imposes an operational burden. For example, in the medical domain, confidential data may be managed on servers isolated from external networks. In such settings, standard FL, which assumes continuous client-server communication, can face practical deployment challenges [Imakura and Sakurai, 2024]. Moreover, strongly isolated operations such as air-gap can make multi-round communication a bottleneck in both operational cost and security risk [Guri, 2024].

When communication rounds are the primary constraint, single-round training frameworks are particularly attractive. Data Collaboration (DC) analysis [Imakura and Sakurai, 2020], a non-model-sharing federated learning paradigm, is one such candidate. However, DC analysis lacks a mechanism to cluster clients under non-IID conditions and thus cannot directly support cluster-wise learning. More broadly, most clustered federated learning approaches require multi-round communication for cluster estimation and training, so single-round end-to-end cluster-wise learning remains insufficiently established. This motivates a one-shot clustered federated learning framework that enables cluster-wise learning under non-IID data.

Accordingly, we propose Data Collaboration-based Clustered Federated Learning (DC-CFL), which builds on the DC analysis framework to jointly perform client clustering and model training while retaining a single communication round. Specifically, DC-CFL measures inter-client similarity based on label distributions and partitions clients via hierarchical clustering. It then performs model training within the DC analysis framework separately for each cluster, aiming to improve accuracy under non-IID conditions within a single-round communication constraint. Figure 1 overviews the DC-CFL workflow. To our knowledge, DC-CFL is the first CFL framework that simultaneously achieves single-round communication and non-IID handling. It thereby offers a practical option for collaborative learning in real-world scenarios with strict communication constraints, such as healthcare and finance sectors operating under air-gapped or strongly isolated environments. Our main contributions are as follows:

- Single-round cluster-wise learning: Building on DC analysis, we develop a CFL framework that performs cluster estimation and cluster-wise model training within a single communication round.
- Empirical validation: Through experiments on benchmark datasets, we demonstrate that DC-CFL (i) attains superior accuracy to existing baselines under non-IID conditions and (ii) substantially reduces the number of communication rounds required by multi-round baselines to reach comparable accuracy.

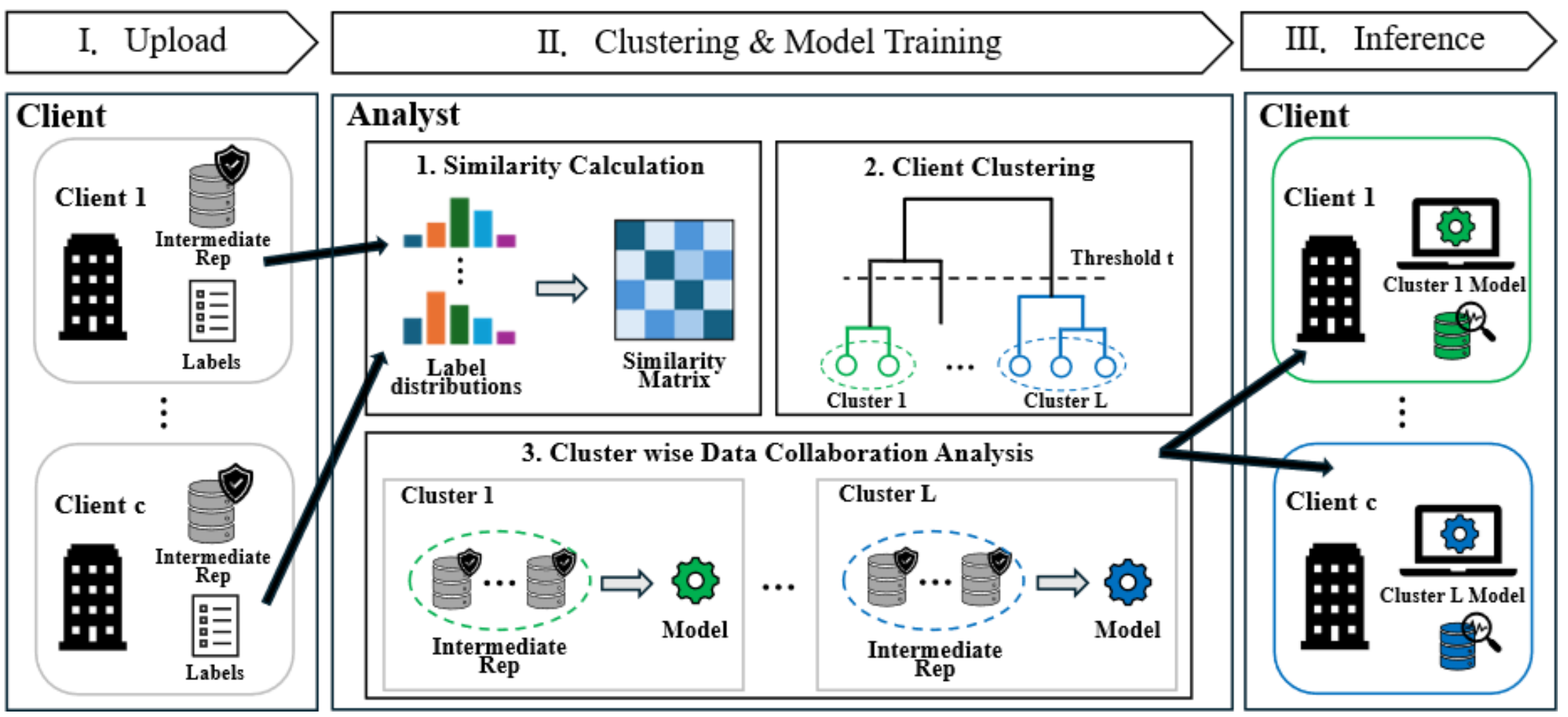

Figure 1: Overview of DC-CFL workflow. Clients upload intermediate representations and labels once; the analyst clusters clients and trains cluster-wise models, then returns client-specific mappings and cluster models for local inference.

## 2 Related Works

### 2.1 FL and Non-IID

Federated Learning (FL) is a framework in which multiple clients collaboratively train a model under the coordination of a central server without sharing raw data [McMahan *et al.*, 2017]. Federated Averaging (FedAvg), a representative FL method, performs client-side local updates and server-side aggregation over multiple communication rounds to optimize a single global model. However, in non-IID conditions where data distributions differ across clients, performance often degrades due to inconsistent update directions (client drift). To address this issue, extensions that still assume a single global model have been proposed, such as FedProx [Li *et al.*, 2020], which introduces a proximal term into FedAvg's local objective to suppress divergence across client updates.

Another line of research on adapting to non-IID data is Personalized Federated Learning (PFL), which aims to obtain a distinct model for each client [Tan *et al.*, 2022]. A representative example is Ditto [Li *et al.*, 2021], which jointly learns a shared model and client-specific models in parallel. Other approaches have also been reported, including methods that adaptively learn aggregation weights over other clients' models [Luo and Wu, 2022] and personalization techniques based on calibration mechanisms [Tang *et al.*, 2024].

However, under severe non-IID conditions, improvements to a single global model often reach a performance ceiling, and PFL can suffer from scalability and overfitting issues due to the growing number of client-specific models and the limited local data available at each client [Sabah *et al.*, 2024]. In such cases, a promising strategy is to group similar clients and perform learning at the cluster level.

### 2.2 Clustered Federated Learning Under Communication-Round Constraints

Clustered Federated Learning (CFL) is a framework that partitions similar clients into clusters and trains models at the cluster level [Sattler *et al.*, 2020]. However, most existing CFL methods rely on multi-round communication to estimate clusters and update models, making communication rounds a dominant operational cost. For example, FedCM [Wang *et al.*, 2025] addresses client clustering and migration based on gradient-path similarity, and DisUE [Leng *et al.*, 2025] integrates dynamic clustering based on parameter similarity with knowledge distillation and an encryption protocol. While these approaches make the problem setting more realistic, they assume multi-round communication during training.

When the number of communication rounds is an operational constraint, one-shot FL (OFL) restricts client–server communication to a single round to reduce communication cost and operational burden [Liu *et al.*, 2025]. However, existing OFL formulations are not specifically designed to realize end-to-end cluster-wise learning under non-IID data. Consequently, enabling CFL to complete both cluster estimation and cluster-wise model training within a single communication round remains insufficiently established under tight communication-round constraints.

To illustrate this gap, FedClust [Islam *et al.*, 2024] is a representative CFL method that estimates clusters via hierarchical clustering without requiring a pre-specified number of clusters, and it has been reported to improve both accuracy and communication efficiency. In FedClust, clustering is performed in a single step, after which cluster-wise training proceeds via standard multi-round federated optimization. This distinction highlights the remaining gap between single-step clustering and fully single-round end-to-end CFL.

# 3 Preliminaries

Data Collaboration (DC) analysis [Imakura and Sakurai, 2020] is a framework for integrated analysis over distributed data without sharing raw data. In DC analysis, each client shares an intermediate representation obtained by dimensionality reduction of its raw data. DC analysis can be positioned as a federated learning approach that does not rely on model sharing [Imakura *et al.*, 2021b]. While conventional FL exchanges model parameters over multiple rounds, DC analysis completes training by sending intermediate representations and labels only once.

In this section, we describe DC analysis for aligning client-specific intermediate representations into a common representation space. Let $c$ denote the number of clients. Let the raw data of client $i$ be $X_i \in \mathbb{R}^{n_i \times m}$, and let the anchor data be $X^{\text{anc}} \in \mathbb{R}^{r \times m}$. The anchor data are provided as shared pseudo-data [Imakura and Sakurai, 2020; Imakura *et al.*, 2023]. Each client $i$ uses an arbitrary dimensionality reduction function

$$f_i: \mathbb{R}^m \to \mathbb{R}^{\widetilde{m}_i}, (0 < \widetilde{m}_i < m), \tag{1}$$

to transform its raw data and the anchor data into intermediate representations:

$$\widetilde{X}_i = f_i(X_i) \in \mathbb{R}^{n_i \times \widetilde{m}_i}, \widetilde{X}_i^{\text{anc}} = f_i(X^{\text{anc}}) \in \mathbb{R}^{r \times \widetilde{m}_i}. \tag{2}$$

Because each client uses a different $f_i$, the intermediate representation $\widetilde{X}_i$ is expressed in a client-specific spaces and cannot be directly concatenated. To align the intermediate representations into a common representation space, we introduce a linear re-transformation $g_i(\widetilde{X}_i) = \widetilde{X}_i G_i$. We estimate $G_i \in \mathbb{R}^{\widetilde{m}_i \times \widehat{m}}$ so that the intermediate representations of the anchor data are aligned to a common representation $Z \in \mathbb{R}^{r \times \widehat{m}}$. Specifically, treating $G_1, \ldots, G_c, and\ Z$ as unknowns, we consider the following least-squares problem:

$$\min_{G_1,\ldots,G_c,Z} \sum_{i=1}^{c} \| Z - \widetilde{X}_i^{anc} G_i \|_F^2 . \tag{3}$$

It has been shown that an approximate solution can be derived via low-rank approximation based on singular value decomposition (SVD) [Imakura and Sakurai, 2020]. First, we compute a low-rank approximation of the matrix obtained by concatenating the anchor-data intermediate representations of all clients along the feature axis:

$$[\widetilde{X}_1^{\text{anc}}, \widetilde{X}_2^{\text{anc}}, \ldots, \widetilde{X}_c^{\text{anc}}] \approx U_1 \Sigma_1 V_1^{\top}. \tag{4}$$

Here, $\Sigma_1$ is a diagonal matrix containing the top singular values, and $U_1\ and\ V_1$ are orthogonal matrices consisting of the leading left and right singular vectors, respectively. We define the common representation as $Z = U_1$; then the transformation matrix for client $i$ is given by:

$$G_i = (\widetilde{X}_i^{anc})^{\dagger} U_1 C. \tag{5}$$

where $(\cdot)^{\dagger}$ denotes the Moore–Penrose pseudoinverse. Assuming an identity matrix for $C \in \mathbb{R}^{\widehat{m} \times \widehat{m}}$, $\widetilde{X}_i^{\text{anc}} G_i$ is aligned with the common representation $Z$ for each $i$. By applying the same transformation $G_i$ to the $\widetilde{X}_i$, we obtain the integrated representation $\widehat{X}_i = \widetilde{X}_i G_i$. Finally, we concatenate $\widehat{X}_i$ along the sample axis to form an integrated matrix $\widehat{X}$, on which we can train an arbitrary supervised learner.

We discuss the privacy implications of our protocol and clarify the associated threat model and assumptions. In this study, clients share neither their raw data $X_i$ nor the dimensionality-reduction functions $f_i$. Instead, the analyst receives intermediate representations together with the corresponding labels only once. By transmitting only intermediate representations and keeping the transformation functions private, the protocol avoids direct sharing of raw data, as in DC analysis [Imakura and Sakurai, 2024]. This protection is derived from the protocol design and the limited invertibility of the transformation, and it is not intended to provide formal guarantees such as differential privacy or cryptographic security. We adopt the standard honest-but-curious analyst model [Imakura *et al.*, 2021a] and focus on deployments in which governance permits sharing intermediate representations and labels. We note that extensions of DC analysis have been proposed, such as adding noise to intermediate representations to satisfy differential privacy and combining dimensionality reduction with noise addition to improve robustness against re-identification attacks [Chen and Omote, 2021; Yamashiro *et al.*, 2024]. Since DC-CFL uses only the same transmitted information as DC analysis and performs cluster estimation solely from the received labels, it requires no additional shared information; accordingly, its threat model and privacy assumptions are equivalent to those of DC analysis.

# 4 Methodology

## 4.1 Overview of DC-CFL

We propose Data Collaboration-based Clustered Federated Learning (DC-CFL), a single-round framework that performs client cluster estimation and cluster-wise learning by extending supervised DC analysis. The procedure consists of three stages (Algorithm 1):

1. Following the DC analysis framework, each client uploads its intermediate representations and the corresponding labels for supervised learning to the analyst only once.

**Algorithm 1 DC-CFL**

**Input**: $\{(X_i, y_i)\}_{i=1}^{c}$, anchor data $X^{anc}$<br>
**Parameter**: threshold $t$<br>
**Output**: $\{h_l\}_{l=1}^{L}$, $\{G_i\}_{i=1}^{c}$<br>
**Dimensionality reduction (Client)**<br>
1: **for** each client $i = 1, \dots, c$ **do**<br>
2: Compute $\tilde{X}_i = f_i(X_i), \tilde{X}_i^{anc} = f_i(X^{anc})$<br>
3: share $\tilde{X}_i, \tilde{X}_i^{anc}, y_i$<br>
4: **end for**<br>
**Similarity Calculation and clustering (Analyst)**<br>
5: Compute label distributions $\{p_i\}_{i=1}^{c}$ from $\{y_i\}_{i=1}^{c}$.<br>
6: Build distance matrix $D$ with $D_{ij} \leftarrow \mathrm{TV}(p_i, p_j)$ $(i \neq j)$.<br>
7: Obtain clusters $\{\mathcal{C}_l\}_{l=1}^{L}$ by Hierarchical Clustering($D;\ t$).<br>
**Cluster-wise Model training (Analyst)**<br>
8: **for** each cluster $l = 1, \dots, L$ **do**<br>
9: Compute $\{G_i\}_{i \in \mathcal{C}_l}$ from $\{\tilde{X}_i^{anc}\}_{i \in \mathcal{C}_l}$ using the DC mapping in Sec.3.<br>
10: Compute $\hat{X}_i = \tilde{X}_i G_i$ for all $i \in \mathcal{C}_l$ and build $\hat{X}_l = [\hat{X}_i]_{i \in \mathcal{C}_l}, Y_l = [y_i]_{i \in \mathcal{C}_l}$.<br>
11: Get $h_l$ $as$ $Y_l \approx h_l(\hat{X}_l)$.<br>
12: **end for**<br>
13: return $\{h_l\}_{l=1}^{L}$, $\{G_i\}_{i=1}^{c}$

2. Using the received labels, the analyst constructs an inter-client distance matrix based on label distributions, performs hierarchical clustering, and determines the final partition.
3. For each cluster, the analyst constructs an integrated representation within the DC analysis framework using only the clients in that cluster and trains an arbitrary downstream model in a cluster-wise manner.

Through these steps, DC-CFL aims to incorporate the benefits of clustering in non-IID conditions.

For cluster estimation in Step 2, we focus on label distribution skew, which has been reported to substantially degrade accuracy in FL [Li *et al*., 2022]. Accordingly, and consistent with prior work demonstrating the effectiveness of clustering based on label distributions [Diao *et al*., 2024; Soltani *et al*., 2023], we define inter-client similarity using the clients' label distributions.

## 4.2 TV-Based Client Clustering

**Client Similarity.** Let $c$ denote the number of clients and let the label set of client $i$ be $Y_i = \{y_{i,1}, \dots, y_{i,n_i}\}$. We consider a $K$-class classification problem where each label takes a value $y_{i,s} \in \{1, \dots, K\}$. We define the label distribution (empirical distribution) of client $i$ as

$$p_i(k) = \frac{1}{n_i} \sum_{s=1}^{n_i} I(y_{i,s} = k), k = 1, \dots, K. \quad (6)$$

Here, $I(\cdot)$ denotes the indicator function. In DC-CFL, we define the inter-client distance $D_{ij}$ by the total variation (TV) distance [Levin *et al*., 2009] between the label distributions $p_i$ $and$ $p_j$:

$$D_{ij} = \frac{1}{2} \sum_{k=1}^{K} | p_i(k) - p_j(k) |. \quad (7)$$

The TV distance is a fundamental measure of discrepancy between two probability distributions, distinct in nature from distances over continuous-space vectors. In DC-CFL, we use the TV distance between empirical label distributions to quantify heterogeneity. This is particularly crucial in one-shot settings, where misclustered clients cannot be corrected. Therefore, we need a criterion that directly penalizes label mismatches to prevent mixing dissimilar clients. Furthermore, clustering based on label skew facilitates the alignment of client-specific representations in supervised DC analysis. Computationally, the TV distance is stable even with zero-frequency classes and requires no smoothing. Since label distributions are represented as probability vectors over a finite class set, TV is theoretically consistent as a comparison measure. For these reasons, we adopt the TV distance and construct the distance matrix $D \in \mathbb{R}^{c \times c}$ using (7). Note that this measure captures only label distribution discrepancies and does not account for feature distribution shifts or concept shifts.

**Hierarchical Clustering.** We partition the clients by applying agglomerative hierarchical clustering [Day and Edelsbrunner, 1984] to the distance matrix $D$. This approach requires only the distance matrix computed from TV distance and does not require the number of clusters to be specified in advance. It constructs a dendrogram in a single run, allowing the granularity of collaboration to be adjusted continuously via a threshold $t$. While various linkage criteria are possible, we adopt complete linkage as a conservative choice to control within-cluster dispersion in one-shot settings, where misclustered clients cannot be corrected via additional rounds. Because complete linkage defines inter-cluster distance as the farthest-point distance, cutting the dendrogram at threshold $t$ yields clusters whose maximum pairwise distance within each cluster (i.e., the cluster diameter) is at most $t$.

## 4.3 Cluster-Wise Model Training

We construct a DC-based integrated representation for each cluster and train a cluster-wise model. Let $\{\mathcal{C}_l\}_{l=1}^{L}$ denote the clusters obtained in Section 4.2. For each cluster $\mathcal{C}_l$, we apply the DC-based integration procedure described in Section 3 to the clients in $\mathcal{C}_l$ and obtain a cluster-specific integrated representation $\hat{X}^{(l)}$. We then train a supervised learner on $\hat{X}^{(l)}$ and its labels to obtain a cluster-wise model $h_l$. Both cluster estimation and cluster-wise learning are performed on the analyst side. Since the analyst already holds all transmitted intermediate representations and labels, hyperparameters can be selected via internal validation splits, requiring no additional client communication. The analyst returns $h_l$ and the mapping matrix $G_i$ to each client $i \in \mathcal{C}_l$ once. For inference, client $i$ computes $\tilde{X}_i = f_i(X_i)$ and $\hat{X}_i = \tilde{X}_i G_i$, and applies $h_l(\hat{X}_i)$.

# 5 Experiments

In this section, we evaluate the effectiveness of the proposed DC-CFL using multiple open datasets. The evaluation focuses on (i) accuracy under non-IID label distributions and (ii) the number of communication rounds required to reach accuracy comparable to DC-CFL. Each client splits its local data into training and test sets and transmits the intermediate representations of the training data and the corresponding labels to the analyst only once. The analyst constructs training and validation sets from the received representations and selects hyperparameters. Final performance is reported as the average test accuracy across clients, where each client evaluates the learned model on its local test set.

## 5.1 Settings

**Datasets.** To evaluate the effectiveness of DC-CFL under realistic data distributions, we conducted experiments using multiple open datasets. Specifically, we used the covertype dataset from Kaggle (we used the entire preprocessed training dataset in which the class frequencies are balanced) [Blackard and Dean, 1999], the Dry Bean dataset publicly available from the UCI Machine Learning Repository [Dry Bean, 2020], and the Gesture Phase Segmentation dataset (hereafter abbreviated as Gesture) [Madeo *et al*., 2013] obtained via OpenML [Vanschoren *et al*., 2014]. For each dataset, Table 1 summarizes the number of samples, the number of features, and the number of label classes.

**Non-IID.** We designed the statistical heterogeneity across clients as label-distribution non-IID, largely following Li *et al*., [2022]. Specifically, we considered the following two types:

- Class-based split (pathological non-IID, C = k): The number of classes held by each client is restricted to $k$. In this section, we use $k \in \{2,3\}$.
- Dirichlet split (Dir($\alpha$)): For each class, the client-wise allocation proportions are generated from a Dirichlet distribution, and $\alpha$ controls the degree of heterogeneity. In this section, we set $\alpha = 0.1$.

In our experiments, we fixed the number of participating clients to 100. This choice was made for a fair comparison, following the experimental setup of FedClust [Islam *et al*., 2024].

**Baselines.** We adopted representative FL/PFL/CFL methods that require multi-round communication as baselines. Specifically, we used FedAvg [McMahan *et al*., 2017] as a standard FL approach, FedProx [Li *et al*., 2020] as a FedAvg variant with a proximal term, Ditto [Li *et al*., 2021] as a representative PFL method, and FedClust [Islam *et al*., 2024] as a representative CFL method. In addition, to examine whether the performance gains of DC-CFL are attributable to cluster-wise learning, we also report results for DC without cluster-specific models and local-only training at each client (Local). Relatedly, one-shot FL (OFL) also targets single-round communication [Liu *et al*., 2025]. However, OFL studies encompass a broad range of protocol assumptions and design choices, and reported performance can be sensitive to these choices. Because our goal is to isolate and quantify the effect of clustering within the DC-based protocol studied in this work, we report representative FL/PFL/CFL methods as widely recognized reference baselines under a limited number of communication rounds and leave a systematic comparison with OFL-specific designs to future work.

| Dataset | samples | features | class |
|---|---|---|---|
| covertype | 15.12K | 54 | 7 |
| Dry Bean | 13.61K | 16 | 7 |
| Gesture | 9.87K | 32 | 5 |

Table 1: Summary of the datasets used in our experiments.

**Accuracy Comparison.** DC-CFL completes training in a single communication round, whereas multi-round baselines require multiple rounds. To approximately match the computational budget, we align the total number of training epochs as the default setting. Specifically, following Bogdanova *et al*. [2020], we set the number of epochs to 50 for Local / DC / DC-CFL, and to 10 epochs × 5 rounds for FedAvg / FedProx / Ditto / FedClust. Note that DC-based methods share representations, whereas standard FL baselines communicate model updates. Therefore cross-paradigm comparisons are provided for reference under tight communication-round constraints and should not be interpreted as a fully matched-protocol comparison.

**Communication-Round Comparison.** To explicitly assess communication efficiency, we increase the number of rounds for multi-round baselines and measure how many rounds are required to reach the accuracy achieved by DC-CFL. In this axis, we do not attempt to match total computation across methods. Instead, we fix the number of local epochs per round for multi-round baselines to 10. We then measure the number of communication rounds required to match the accuracy of DC-CFL, which completes training in a single round with 50 training epochs. We conduct the round comparison only for label-skew settings with $C \in \{2,3\}$ under the class-based split. In this split, each client's class subset is explicitly controlled, making the required rounds easier to interpret as differences in convergence behavior under heterogeneity. In contrast, the Dirichlet split depends on random draws and can exhibit substantial across-trial variance, making the required-round comparison unstable. Accordingly, we report accuracy comparisons for all settings, while limiting round-comparison results to the class-based split.

**Implementation details.** To ensure reproducibility, we summarize the key implementation settings below. All experiments were conducted using Python 3.10 and PyTorch 2.9 in a standard CPU environment.

- Dimensionality reduction: For the dimensionality-reduction function $f_i$, we adopted Principal Component Analysis (PCA) [Pearson, 1901], which exhibited the most stable and highest performance in our preliminary experiments compared with alternatives such as Locality Preserving Projection (LPP) [He and Niyogi, 2003] and Random Projection (RP) [Johnson and Lindenstrauss, 1984]. We set the intermediate representation dimension to $\widetilde{m} = m - 1$.

| | covertype | | | Dry Bean | | | Gesture | | |
|---|---|---|---|---|---|---|---|---|---|
| Method | C = 2 | C = 3 | Dir(0.1) | C = 2 | C = 3 | Dir(0.1) | C = 2 | C = 3 | Dir(0.1) |
| Local | 0.8955 | 0.7611 | 0.8515 | 0.9805 | 0.9452 | 0.9288 | 0.7033 | 0.5304 | 0.8202 |
| DC | 0.7753 | 0.7613 | 0.7743 | 0.9156 | 0.9136 | 0.9090 | 0.5171 | 0.5254 | 0.5180 |
| FedAvg | 0.4005 | 0.5048 | 0.3696 | 0.3286 | 0.5156 | 0.3036 | 0.3380 | 0.4094 | 0.2859 |
| FedProx | 0.4172 | 0.5262 | 0.4249 | 0.3352 | 0.5191 | 0.3233 | 0.3390 | 0.4379 | 0.2983 |
| Ditto | 0.8970 | 0.7575 | 0.8463 | 0.9715 | 0.9124 | 0.9106 | 0.7038 | 0.5326 | 0.8230 |
| FedClust | 0.9040 | 0.7635 | 0.8474 | 0.9722 | 0.9156 | 0.9093 | 0.7058 | 0.5339 | 0.8227 |
| **DC-CFL** | **0.9382** | **0.8389** | **0.8618** | **0.9861** | **0.9672** | **0.9371** | **0.7557** | **0.6162** | **0.8256** |

Table 2: Accuracy results. Best in each setting is in **bold**. (FL baselines are reported as reference due to protocol differences)

- Anchor data: Following Imakura *et al.*, [2023], we generated 1,000 samples.
- Dimension of the integrated representation: In the SVD applied to the concatenation matrix of the anchor-data intermediate representations, preliminary experiments indicated that using components with singular values no smaller than $10^{-2}$ yielded the best performance; we therefore determined the dimension of the integrated representation accordingly.
- Model: For fair comparison, we used the same two-layer MLP architecture for all methods. Specifically, the input dimension was matched to each method's input representation (raw features or the integrated representation), the hidden-layer sizes were set to either [64, 32] or [32, 16], the output dimension was set to the number of classes, and ReLU was used as the activation function.
- Hyperparameters: We used SGD for optimization with a batch size of 32. We adopted the SGD-based optimizer settings in Islam *et al.*, [2024] as a reference and matched the learning rate and momentum settings where applicable. Specifically, we fixed the learning rate to 0.01 and the momentum to 0.5 for DC, Local, DC-CFL, Ditto, and FedClust. For FedAvg and FedProx, given their sensitivity in convergence behavior, we selected the learning rate $\eta \in \{0.1, 0.01, 0.001\}$ based on validation performance and set the momentum to 0.9. The FedProx proximal-term coefficient was selected from $\mu \in \{1, 0.1, 0.01, 0.001\}$, and the Ditto regularization coefficient was selected from $\lambda \in \{2, 1, 0.1, 0.01\}$, also based on validation performance. Moreover, the clustering distance threshold $t$ in DC-CFL and FedClust was selected using validation data from the candidate set $\mathcal{T} = \{0.1, 0.2, \ldots, 0.9\}$. Notably, DC-CFL does not require additional communication for selecting $t$. We set the client participation rate to 100% in all communication rounds.
- Number of runs: We conducted 10 independent runs with different random seeds and report the mean results.

## 5.2 Results and Discussion

**Accuracy**

As shown in Table 2, under the default setting, DC-CFL achieves higher accuracy than the existing methods. DC-CFL outperforming Local indicates that integrating information across clients can be beneficial even when raw data cannot be shared. Moreover, DC-CFL's outperformance of conventional DC analysis suggests that, under label distribution skew, clustering clients with similar label distributions and performing DC-based integration and learning per cluster can be effective. Since heterogeneity in this paper is primarily induced by label skew, clustering clients with similar label distributions is expected to reduce discrepancies among client-specific representations that DC analysis needs to align. This can reduce alignment errors in constructing the integrated representation and improve downstream performance.

**Single-Round Advantage**

Compared with multi-round baselines, DC-CFL substantially reduces the number of communication rounds needed to reach DC-CFL's accuracy level (Table 3). FedAvg, FedProx, and Ditto do not reach this target within $R \leq 50$ under our hyperparameter settings. This behavior may be attributable to their sensitivity to client drift under strong non-IID conditions (FedAvg/FedProx) and to the reliance on global updates, which can cause performance to plateau under severe heterogeneity (Ditto). By contrast, FedClust improves as the number of rounds increases and reaches a level comparable to DC-CFL in 19–33 rounds for C=2 and 15–27 rounds for C=3, depending on the dataset. Figure 2 shows an example of accuracy trajectories on covertype (C=3), illustrating that FedClust benefits from additional rounds while the other baselines fail to reach the same level within $R \leq 50$.

| | covertype | | Dry Bean | | Gesture | |
|---|---|---|---|---|---|---|
| **Method** | C=2 | C=3 | C=2 | C=3 | C=2 | C=3 |
| **FedClust** | 24 | 15 | 19 | 16 | 33 | 27 |

Table 3: Number of communication rounds required by each method to match or exceed the accuracy of DC-CFL.

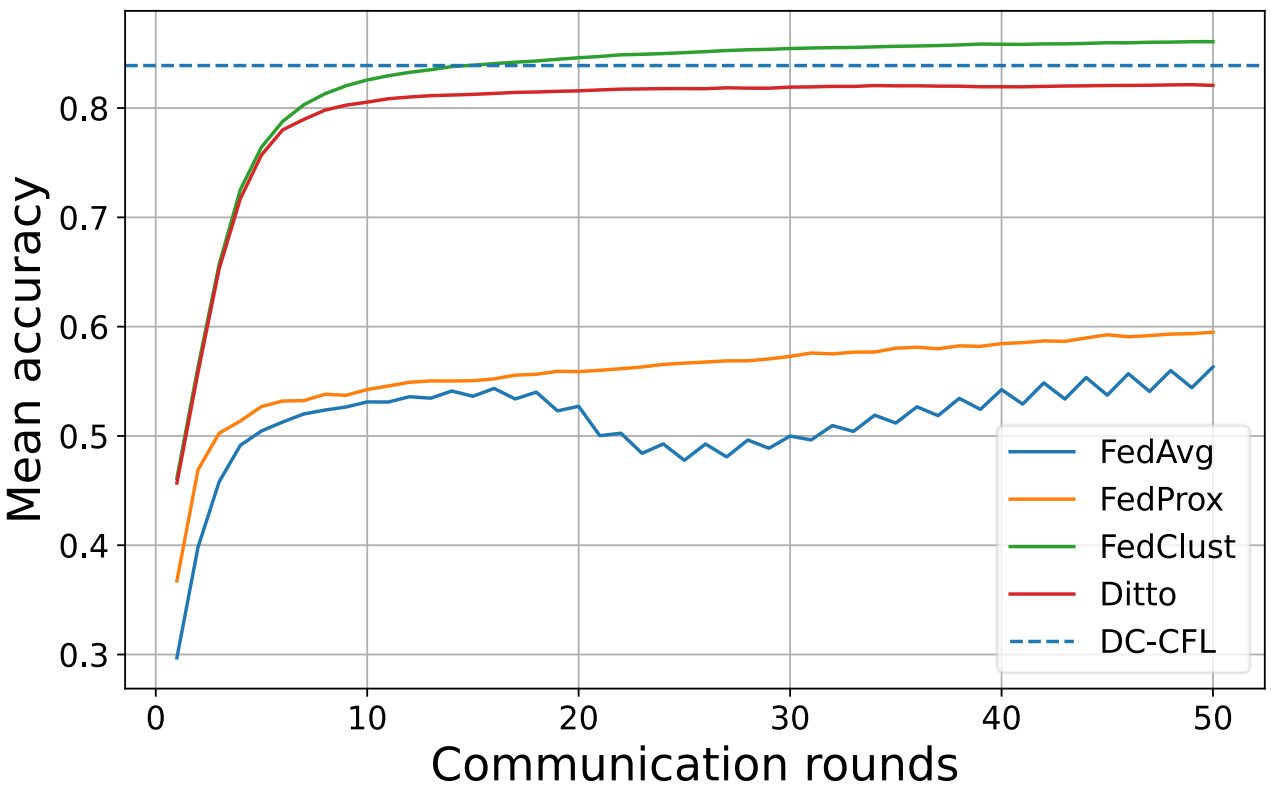

Figure 2: Mean test accuracy versus communication rounds on the covertype dataset under the non-IID setting (C = 3).

Overall, DC-CFL is advantageous when the number of communication rounds is the primary constraint, whereas multi-round CFL methods can be effective when sufficient rounds are permissible. To complement the round-based comparison, we also estimated the per-client total communication volume on the covertype setting. In DC-CFL, per-client communication is dominated by a single upload of intermediate representations $\tilde{X}_i \in \mathbb{R}^{n_i \times \tilde{m}_i}$ and $\tilde{X}_i^{\mathrm{anc}} \in \mathbb{R}^{r \times \tilde{m}_i}$ (with labels), and a single download of the mapping $G_i \in \mathbb{R}^{\tilde{m}_i \times \hat{m}_l}$ and the cluster-wise model $h_l$, totaling approximately 0.27 MB per client. In contrast, FedClust requires 15–24 rounds (approximately 0.70–1.12 MB) to reach comparable accuracy, far exceeding the total communication volume of DC-CFL. Note that the communication volume of DC-CFL depends on key design choices: the number of anchor data samples $r$, the dimension of the intermediate representation $\tilde{m}_i$, and the dimension of the cluster-wise integrated representation $\hat{m}_l$.

**Sensitivity Analysis of Distance Threshold $t$**

We conduct a sensitivity analysis of the clustering distance threshold $t$ using the covertype dataset under two non-IID conditions (C=2 and C=3) to evaluate its impact on DC-CFL performance under different levels of data heterogeneity. As shown in Figure 3, $t$ is a sensitive hyperparameter, and its optimal value depends on the degree of heterogeneity. These results suggest that tuning $t$ for each heterogeneity setting, rather than using a fixed value, can systematically improve accuracy, making it an effective control parameter.

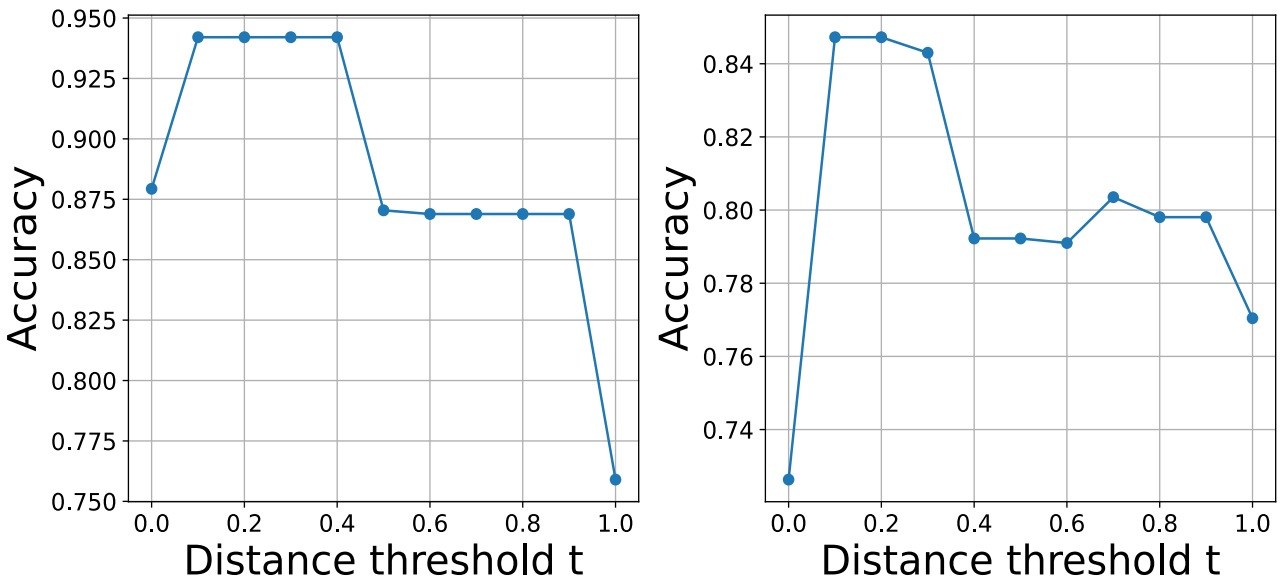

Figure 3: Accuracy of DC-CFL under varying distance threshold $t$ on the covertype (Left: C=2, Right: C=3).

# 6 Conclusion

In this study, building on DC analysis, we proposed Data Collaboration-based Clustered Federated Learning (DC-CFL), which completes both cluster estimation and cluster-wise model training within a single communication round. Experiments on open datasets demonstrate that, under representative non-IID conditions, DC-CFL achieves accuracy comparable to that of existing methods that require multiple communication rounds. These results indicate that DC-CFL can be a practical option for collaborative AI model development even in environments where confidential data are distributed across organizations and the number of communication rounds is severely constrained.

As future work, we plan to address the following issues. First, we will investigate automatic selection of the distance threshold $t$ for hierarchical clustering. Second, through a systematic comparison with one-shot federated learning (OFL) methods, we aim to clarify the position of DC-CFL in the trade-off among accuracy, communication cost, and information disclosure risk. Third, it is necessary to extend the evaluation of the robustness and security of both cluster estimation and learning outcomes under a broader range of non-IID conditions and adversary models.

Notably, the proposed method is designed under the assumptions described in Section 3 and does not provide formal guarantees such as differential privacy. In addition, in prioritizing single-round communication, this paper assumes that sample-level label sharing required for supervised DC analysis is permissible; however, this assumption may limit the applicability of the method. In future work, we will quantitatively evaluate the privacy–accuracy trade-off, including the introduction of differential privacy to the transmitted information. Furthermore, we will explore designs that minimize the information to be shared, for example via aggregation or secure and hidden computation. In particular, since clustering in our framework can be performed using label distributions, we aim to avoid sharing sample-level labels required for supervised DC analysis. We expect these extensions to make DC-CFL both more broadly applicable and more explicit about its privacy properties.

## Acknowledgments

This work was supported by JSPS KAKENHI Grant Number JP23K22166.